\documentclass{article}

\usepackage{arxiv}

\usepackage[utf8]{inputenc} 
\usepackage[T1]{fontenc}    
\usepackage{hyperref}       
\usepackage{url}            
\usepackage{booktabs}       
\usepackage{amsfonts}       
\usepackage{nicefrac}       
\usepackage{microtype}      
\usepackage{lipsum}		
\usepackage{graphicx}
\usepackage{natbib}
\usepackage{doi}
\usepackage{titletoc}

\usepackage{gensymb}
\usepackage{adjustbox}
\usepackage{epsfig}
\usepackage{caption}
\usepackage{multirow}
\usepackage{subcaption}

\title{Deep Active Learning in the Presence of Label Noise: A Survey}

\author{ \href{https://orcid.org/0000-0002-9191-0565}{\includegraphics[scale=0.06]{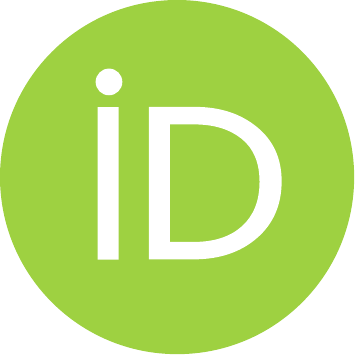}\hspace{1mm}Moseli Mots'oehli}\\
	Department of Information and Computer Science\\
	University of Hawai'i at Manoa\\
	Honolulu, HI 96822 \\
	\texttt{moselim@hawaii.edu} \\
        \And
	\href{https://orcid.org/0000-0000-0000-0000}{\includegraphics[scale=0.06]{orcid.pdf}\hspace{1mm}Kyungim Baek} \\
	Department of Information and Computer Science\\
	University of Hawai'i at Manoa\\
	Honolulu, HI 96822 \\
	\texttt{kyungim@hawaii.edu} \\
}

\renewcommand{\headeright}{A Survey}
\renewcommand{\shorttitle}{Active Learning With Label Noise}

\hypersetup{
pdftitle={Deep Active Learning in the Presence of Label Noise: A Survey},
pdfsubject={q-bio.NC, q-bio.QM},
pdfauthor={Moseli Mots'oehli},
pdfkeywords={ Deep Active Learning, Label Noise, Image Classification, Vision Transformer},
}

\begin{document}
\maketitle

\begin{abstract}
Deep active learning has emerged as a powerful tool for training deep learning models within a predefined labeling budget. These models have achieved performances comparable to those trained in an offline setting. However, deep active learning faces substantial issues when dealing with classification datasets containing noisy labels. In this literature review, we discuss the current state of deep active learning in the presence of label noise, highlighting unique approaches, their strengths, and weaknesses. With the recent success of vision transformers in image classification tasks, we provide a brief overview and consider how the transformer layers and attention mechanisms can be used to enhance diversity, importance, and uncertainty-based selection in queries sent to an oracle for labeling. We further propose exploring contrastive learning methods to derive good image representations that can aid in selecting high-value samples for labeling in an active learning setting. We also highlight the need for creating unified benchmarks and standardized datasets for deep active learning in the presence of label noise for image classification to promote the reproducibility of research. The review concludes by suggesting avenues for future research in this area.
\end{abstract}


\section{Introduction}\label{sec:intro}
Machine learning algorithms are a sub-class of artificial intelligence that learns from data to perform a pre-defined task such as classification, regression, or clustering. Of the numerous algorithms for machine learning, artificial neural networks, deep neural networks in particular have done exceptionally well in tasks involving complex data representations such as images, text, and sound. The main reason for this is that if you have a large enough dataset, you can build more extensive and complex models with little to no risk of over-fitting. While this works in theory, the practical applications have major drawbacks such as the need for labeled training examples that come at a high cost due to the time needed to label the data, the high cost of labor in very specialized fields, or the cost of running simulations that would produce the ground truth dataset. The solution comes in the form of deep active learning (DAL) algorithms, which strive to let the learning algorithm iteratively pick data examples to be labeled from a larger unlabelled dataset, in such a manner that results in: (1) A smaller labeled training set, (2) A dataset that is representative of the underlying data distribution leading to a near-optimal learner, (3) A data labeling skim that does not exceed the labeling budget. 

While this works well for most use cases, real-world dataset labeling has inherent label noise due to a variety of factors such as redundant observations being labeled differently, the best human expert classification performance being low, or the use of auto-labeling software such as Mechanical Turk. This has adverse effects on these DAL algorithms' performance, and most existing DAL literature focuses on noise-free settings. We explore existing literature around the problem of using DAL algorithms in the existence of label noise. We are particularly interested in the image classification domain using different deep representation learning frameworks such as convolutional neural networks (CNNs) and vision transformer networks. 

In Section \ref{sec:deep_learning}, we briefly discuss deep learning and the architectures used in image classification. Section \ref{sec:Activelearning}, presents the main ideas behind active learning as well as the issues that arise when datasets contain noisy labels. In Section \ref{sec:EvaluationAndDatasets}, we detail commonly used datasets for active learning on image classification tasks as well as the evaluation metrics. Section \ref{sec:DALAFNLabels} is a detailed analysis of the literature on active learning with label noise in image classification tasks. We conclude by exploring possible directions for future research in DAL on vision tasks under label noise.



\section{Deep Learning}\label{sec:deep_learning}
Deep Learning (DL) refers to the use of artificial neural networks (ANNs) with multiple hidden layers \cite{ivakhnenko:GMDH65}, to approximate known or unknown functions. The multi-layered neural network was built on top of the perceptron \cite{Rosenblatt:Perceptron58} introduced in 1958. Over the years, different domain-specific DL architectures have been developed to enhance the quality of the learned representations from the different data modalities. Early research focused on improving optimization, custom layers and connections, activation functions, loss functions, and hyper-parameter tuning techniques for the multi-layer perceptron as a way to improve performance on different data modalities. For tabular data, tree-based ensemble learning algorithms such as Random forest \cite{Breiman:RandForest01}, XGBoost \cite{Roman:XGBoost16}, and CatBoost \cite{Prokhorenkova:CatBoost18} are preferred over DL for their superior performance and resource efficiency. A non-exhaustive selection of interesting neural network adaptations to tabular data includes \cite{Schafl:Hopula22,Roman:Tabulartransfer22,Popov:NobviousDLTabular20,Arik:TabNet21,Baohua:SuperTML19}. In the natural language processing domain, earlier work involved learning word and sentence representation using shallow neural networks in an unsupervised setting \cite{pennington:Glove14,novak:cbow20,grave:Fasttext18}. Until the wide adoption of attention-based transformer language models \cite{Vaswani:RealAttention17,See:Attention17}, word and sentence level embeddings are fed to a DL model with recurrent connections such as a Long-Short-Term-Memory(LSTM) network \cite{Hochreiter:LSTM97} to achieve state-of-the-art results on down-stream text classification, sentence completion, named entity recognition or summarization tasks. For non-temporal visual tasks such as image classification, object detection, segmentation, and pose estimation \cite{Artacho:UniPose20}, CNN-based architectures with specialized output layers and a lot of training data are still the most widely adopted approach. With each of these complex tasks, there are different challenges in the data annotation process that introduce varying levels of label noise.

While the DL methods discussed in this section have been applied  to other supervised learning vision tasks such as detection and segmentation, we focus on approaches for image classification in this section. We give a brief overview of CNNs that are responsible for a large share of progress in vision-based tasks. We then highlight the use of more complex CNNs for image classification and finally explore the literature on state-of-the-art spatial attention-based models (Vision Transformers) in the context of image classification \cite{Kolesnikov:ViT21}. 

\subsection{Convolutional Neural Networks}\label{sec:cnn} 
Convolutional neural networks were introduced by Yann Lecun and Yashua Bengio as an improvement to human-based feature extraction in training multi-layer neural networks on spatial data \cite{Lecun:lecunGradientBased98}. The key deficiencies with training fully connected feed-forward neural networks (FFNN) using back-propagation for computer vision tasks are efficiency and transformation (rotation, translation) invariance. Handling high-dimensional image data with standard input neurons is non-trivial and inefficient. Given that low-resolution image datasets are normally $28\times 28$, the initial input layer using an FFNN would contain $784$ neurons. If the subsequent hidden layer had as little as $100$ neurons, the 2-layer fully connected network immediately has more than $78400$ weights and bias terms connecting the two layers. The weights of the network are stored in high-dimensional matrices, and the flow of information in the forward and backward pass is performed using matrix operations. The number of input neurons and hidden layer depth required for accurate approximation of complex image-to-class mappings on high-resolution images using FFNNs is large.

\begin{figure}[!tbp]
	\begin{center}
		\includegraphics[width=0.75\columnwidth]{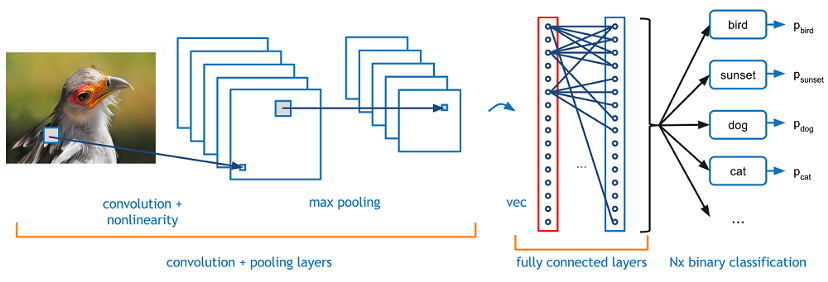}
	\end{center}
\caption{[Source: \href{https://towardsdatascience.com/covolutional-neural-network-cb0883dd6529}{Standard CNN}]. CNN for classifying an image into one of the categories:  bird, sunset, dog, cat, and more common objects.}
\label{fig:SimpleCNN}
\end{figure}

CNNs are especially good at handling image data for three main reasons; firstly the convolution operation uses a sliding filter to identify and highlight the presence of local relations between pixels that represent important features. By so doing, we capture features expressing lines, edges, and corners implicitly. Secondly, in CNNs, the input image is not flattened into an array as is the case in using FFNN. This means the relative positions of pixels in a grid format are preserved and so we do not lose information through rearranging the pixels. Finally, in CNNs, filters have their own weights, but the same filter is used to slide over the image, and this means the features learned are invariant to the positions of patches on the image as the convolution operation learns local relationships between pixels. In addition, this way CNNs are able to use fewer weights than would be the case if we were to consider the absolute positions of pixels on the image and capture positional encoding. Figure \ref{fig:SimpleCNN} depicts a simple CNN with convolution, pooling, and fully connected layers with non-linearity activation functions for image classification. 

\begin{figure}[!tbp]
	\begin{center}
		\includegraphics[width=0.75\columnwidth]{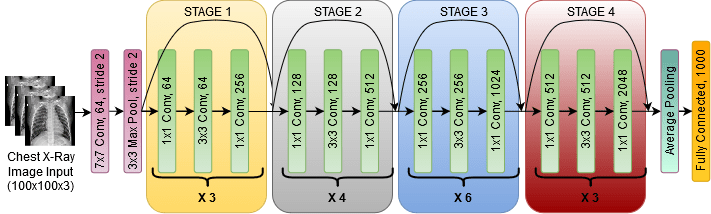}
	\end{center}
\caption{Resnet50 architecture with convolutional blocks of different filter sizes and max pooling. Residual connections acting as memory cells arch above the blocks passing initial information all the way to the final layer \cite{He:ResNetL16}.}
\label{fig:ResNet50}
\end{figure}

More advanced CNN architectures have been introduced, mostly similar in that they have multiple convolution and pooling blocks (earlier layers capture low-level features, and deeper layers capture higher-level features). The most notable of these are GoogLeNet \cite{Szegedy:GoogLeNet15}, VGG \cite{Simonyan:VDCNN15}, ResNet \cite{He:ResNetL16},  DenseNet \cite{Huang:DensNet17}, and EfficientNet \cite{Tan:EfficientNet19}. For example, ResNet, as shown in Figure \ref{fig:ResNet50}, demonstrated the idea of residual connections (also called skip connections). The residual connections pass one layer's inputs directly to the next convolution block to provide lower-level context to the subsequent layer hence combating vanishing gradients in very deep networks. DenseNet on the other hand has a dense building block in that all the layers in a block have direct connections with each other, allowing for a more effective reuse of features in the network. Also, by having all layers connected, a regularization effect is created so that the network does not learn redundant representations, hence combating over-fitting. 

The different layers are connected by non-linear activation functions such as the popular ReLU and Elu \cite{Nair:ReLu10,Clevert:ELu16}. CNNs have been the dominant approach to computer vision benchmarks for a large part of the last decade mainly due to their ability to extract meaningful spatial features from images. The main catalysts in ascending order of importance for this were the availability of large labeled training datasets, advances in computing hardware, and a reduction in the computational cost of training such Deep Neural Networks (DNNs). Post ImageNet \cite{Jia:ImageNet09}, CNN-based models trained on very large labeled datasets have been used in the feature extraction and pre-training step of most fine-tuned state-of-the-art approaches in different vision tasks.

\subsection{Vision Transformers}\label{sec:vision_transformer}
Before full transformer models in the language domain, the best LSTM models use a low dimensional vector representation to pass information from an encoder network to a decoder network, while using an attention mechanism. Attention in this setting is used to learn what parts of an input sequence are most important in predicting different parts of the output. In the original paper "Attention is all you need" \cite{Vaswani:RealAttention17}, Vaswani et al. demonstrate that long temporal dependencies can be learned without the need for recurrence. The three fundamental components in a transformer network are a positional encoding of words, attention, and self-attention mechanisms. Positional encoding of both input and output tokens is achieved by assigning integer values to tokens/words based on their relative position in the input and output sequences. Unlike LSTMs, the work of learning word progression and relationships between input and output words is done implicitly by the network instead of designing networks with explicit bias in the form of recurrent cells and sequential processing. Self-attention makes it possible to learn good representations for most languages given a sufficiently large collection of text in a semi-supervised manner by masking tokens and letting the network learn what the missing word is in any given input sequence. The learned representations are then used on a downstream task with fewer labeled data. Because transformers do not process input tokens in sequence, they are perfect for parallel GPU training.

\begin{figure}[!tbp]
	\begin{center}
		\includegraphics[width=0.75\columnwidth]{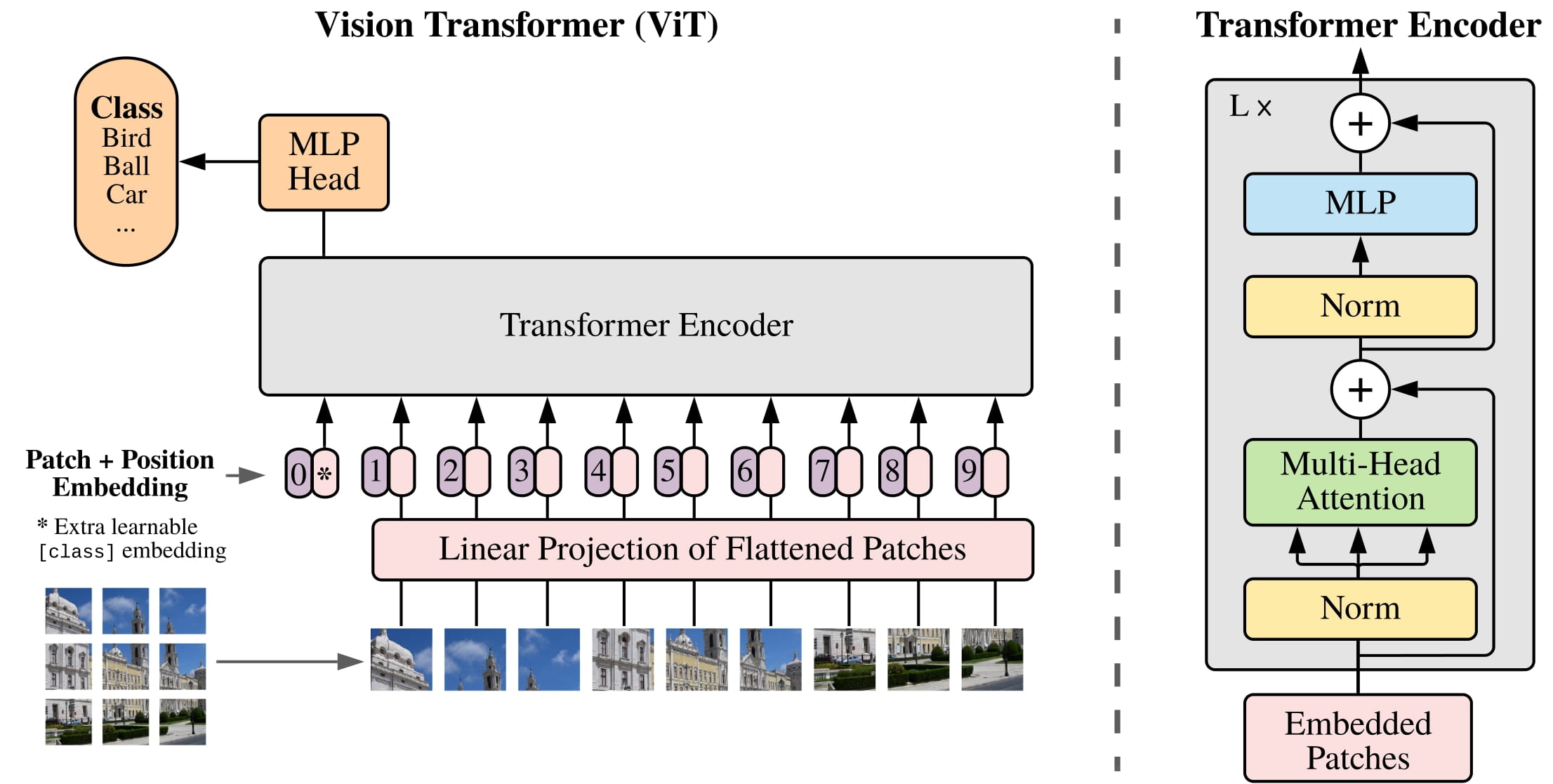}
	\end{center}
\caption{Vision transformer architecture showing an input image split into 14 by 14 patches, and linearly projected to the standard transformer input space. The far right side of the image shows the components of the standard transformer block with multi-head attention \cite{Kolesnikov:ViT21}.}
\label{fig:ViT}
\end{figure}

Like most great innovations, the fundamental ideas of the transformer have been incorporated into CNNs \cite{wortsman:ModelSoups22,Zihang:CoAtNet21,Srinivas1:Bottleneck15}, and in some cases completely replacing CNNs \cite{touvron:deeperwithViT20,liu:swin21,Chen:PaLI22} to produce state-of-the-art results in various computer vision benchmarks. In \cite{Kolesnikov:ViT21}, Kolesnikov et al. present the earliest vision transformer(ViT) to surpass state-of-the-art CNNs on most image classification benchmarks. They show that in the large dataset regime, ViTs achieve higher classification accuracy, are more computationally efficient, and show no signs of saturation compared to CNNs such as ResNet and EfficientNet on increasingly larger datasets. The main difference between the natural language processing (NLP) transformers and the vision transformers is in how the input is encoded. With vision transformers, they take 14 by 14 patches from an image, flatten them, and apply a linear projection onto a higher dimensional space equal to that of the original input space of the NLP transformer. The spatial proximity relations of patches are implicitly left to the transformer to learn in the following way: They add a trainable 1D positional encoding vector to each patch's linear projection. The positional representations are organized in the order of the patches starting from the top left corner to the bottom right corner of the image as depicted in Figure \ref{fig:ViT}. Beyond input encoding, the rest of the ViT architecture is similar to that of the language transformer for classification tasks.

The paper shows interestingly that, through the attention mechanism the transformer layers are able to learn the same low-to-high level features with increasing depth as is the case with deep CNNs. Other notable implementations of ViTs for image classification without label noise include \cite{Yu:CoCa22,Chen:PaLI22,liu:swin21}. ViTs are included in this review and further discussed in Section \ref{sec:Conclusion} as we perceive them to be a very important area for future research. This is because they are progressively becoming the dominant multi-modal approach and yet very little work has been done in applying them to DAL and learning with label noise for image classification. These models are designed in a modular fashion to easily be able to learn both language and image representations for image captioning, classification, scene-text understanding, and visual question answering \cite{Yu:CoCa22}. It is particularly interesting since the authors present a joint contrastive loss (image-to-text and text-to-image), image classification loss, and image-to-language captioning loss, allowing for efficient training of a single network for multiple tasks, and the ability to transfer the learned representations to a different downstream task and dataset.

In the next section the active learning (AL) framework for machine learning is described, including key approaches for training deep learning models on a labeling budget in the case of clean labels, and finally, the scene is set for label noise and the literature addressing DL on noisy labels.

\section{Active learning}\label{sec:Activelearning}
\begin{figure}[!tbp]
	\begin{center}
		\includegraphics[width=0.75\columnwidth]{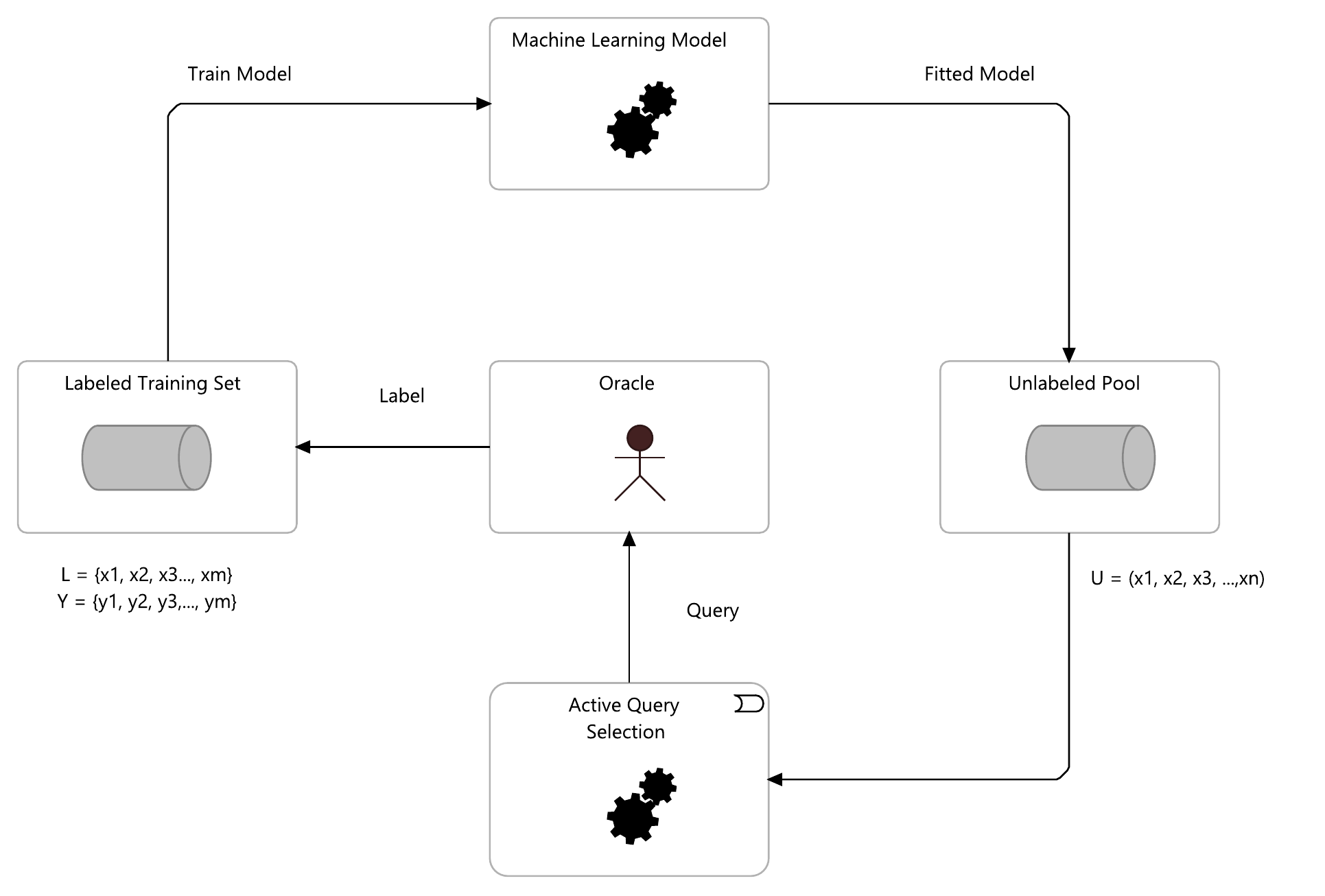}
	\end{center}
\caption{The five main components to the standard Active Learning Framework. Each of these components may vary depending on the complexity of the data to be learned and the available resources. Most work in active learning has focused on the development of query selection algorithms that lead to highly informative and diverse data samples for labeling by the oracle.}
\label{fig:ALFramework}
\end{figure}

In most supervised machine learning use cases, there is an initial data collection and labeling cost, in both money and time. In some domains and tasks, datasets are inherently difficult to label for a variety of reasons, meaning more time is needed even by an expert human annotator to assign a label to each sample. In other cases the cost of hiring expert annotators is high, such as is the case in medical imaging \cite{Grriz:CostEffectiveA17,Konyushkova:LearningAL17}, or the cost of producing the samples is high, such as is the case in experimental physics where observations come from very expensive telescopes or particle accelerators. This presents a challenge to the real-world use of machine learning systems, especially as unlabeled dataset sizes increase. Active learning is a machine learning paradigm, as depicted in Figure \ref{fig:ALFramework}, that seeks to address this problem by letting learning algorithms iteratively select a subset $L^{m}$ of size $m$, from a larger unlabelled dataset $U^n$ of size $n: m \leq n$, to be labeled by an oracle $O$ for training. The active learning mantra can be stated as follows: Train a machine learning model on a significantly smaller labeled dataset, with little to no drop in test performance, all the while staying within a pre-determined labeling budget $B$.

Deep active learning algorithms (DAL), while overlapping, can broadly be grouped into pool-based methods, density-based methods, and data expansion methods. Pool-based methods select samples for labeling from an unlabeled pool, based on either the uncertainty of the currently trained model on samples $U^{n}$, the diversity of samples in the labeled set $L^m$ used to train the current model or a combination of both \cite{Lewis:PoolBased94,McCallum:EmployingEA98,Shui:DeepUnified20}. Pool-based methods are simple in their formation and implementation but can be computationally expensive for large datasets of high dimensional data such as images. Since pool-based methods largely rely on metrics evaluated on the entire unlabeled dataset to select new candidates, this is not ideal for applications that require low latency. Density-based methods seek to capture key characteristics of the underlying data distribution. This is done by selecting a core-set of samples for labeling that are sufficiently representative of the entire dataset, and leads to good generalization \cite{Sener:CoreSetActiveL18,Phillips:NearCoreSets,Phillips:CoreSetsSketches16}. More recent literature blends pool and density-based methods to take advantage of each approach's benefits. These methods thus lead to efficient and robust models trained on core-sets containing diverse samples that maximize the margins between object classes \cite{HarPeled:MaxMarginCoreSet07,Geifman:DeepALtail17}. Some methods in this approach use the hidden layer representations from training a self-supervised task on the image data, instead of the raw pixels. These include pre-training on image orientation, random ($90,180,270,360)\degree$ rotation classification, or self-supervised contrastive learning, where the target is an arbitrary patch of adjacent pixels in the image \cite{Chen:ContrastiveVR20,Du:ContrastiveMismatch21,Wang:Contrastive21}. Data expansion methods seek to expand the training dataset, by generating reasonably realistic synthetic data samples for each target class to enhance the learning algorithm's performance on the real test dataset \cite{Chen:ContrastiveVR20}. Since their introduction, Generative Adversarial Networks (GANs) and their variations \cite{Goodfellow:GANs14,Gonog:GANReview19,Sinha:VAEAL19} were the go-to method for generating synthetic data. However, the training of GANs is unstable, the samples tend to be unrealistic, and it is hard to evaluate these samples for quality \cite{Barnett:GANConvergence18,Mescheder:GANConvergance18}.

\subsection{Label Noise}\label{subsec:LabelNoise}
label noise refers to the scenario in which data labels are corrupted, with or without intention, so that we do not have $100\%$ confidence in their correctness. Label noise is different from feature noise which is normally used to refer to adding gaussian noise to feature values. Label noise impacts learning algorithms more adversely than feature noise does, and is harder to deal with \cite{Chicheng:ActiveStrongWeakLabelers15,Wei:NoisyAnnotation22,Algan:ImageNoisySurvey,Cordeiro:DeepNoisyAnnotations20}. Label noise is inherent in the data collection and processing life-cycle. Most real-world datasets are subjected to a number of label noise sources based on how the data is collected, curated, and stored. Label noise in practice broadly stems from (1) incorrect crowd-sourced labels where the annotators are non-experts such as is the case with \cite{ClickWorker}, and \cite{MechanicalTurk}, (2) incorrect expert annotations due to the complexity of the data, as is common in medical fields \cite{Grriz:CostEffectiveA17}, (3) labeling errors introduced by automatic labeling by web crawling software and other AI labeling systems such as \cite{ScaleAI}, (4) noise introduced by multiple experts or non-experts labeling the same sample differently.

Learning noisy labels is especially hard due to the fact that cost functions are generally significantly less complex than feature extraction layers. Label noise can be grouped, and is mostly treated based on what is known about the noise-generating distribution \cite{Nagarajan:LearningNoisy13}. Some datasets contain label noise from a known and quantifiable generative distribution, while in other cases, too little or nothing is known about the noise transition matrix to model. Label noise can be class-independent or class-dependent. Class-independent label noise is the easiest to generate. The generative process can be summarized in this manner: for each sample, the class label is replaced with a random class label, with a fixed probability $1/N$ where $N$ is the number of classes \cite{Patrinin:LossCorection17}.
Class-dependent label noise is normally a result of expert human annotation. It results from pairs of very closely related or indistinguishable classes being occasionally mislabeled \cite{Han:MaskingNoisy18}. For example, the true large-sized cat is occasionally labeled as a small dog, and visa versa. Common methods for training DL models include first filtering out samples with a high probability of being noisy and iteratively training on a dataset with trusted labels until a threshold is reached. The filtering process in most literature involves training two different neural networks with a custom loss, and monitoring samples on which they disagree on predictions. This method works well since it has been shown that the networks train on stronger signals first, which is the case in a dataset with predominantly clean labels. Representative methods in this approach, trained in a non-active learning manner include Decoupling \cite{Malach:Decoupling17} and Co-teaching \cite{Han:Coteaching18}. The main implementation difference between the two approaches is in how the two networks' weights are updated. Decoupling updates each network's weights based on its prediction error when the networks have a prediction disagreement. Co-teaching on the other hand, cross-updates the weights with the error signal from the other network. Unlike Decoupling, Co-teaching addresses noisy labels explicitly by enabling the networks to peek into each other's hidden state, simultaneously reducing the risk of each network over-fitting the noisy input.

We have introduced deep learning, active learning, and learning with label noise.  For further reading, we suggest the survey papers \cite{Algan:ImageNoisySurvey,Ren:DALSurvey20} on image classification with noisy labels, and DAL on clean labels respectively. DAL methods on noisy labels are presented in Section \ref{sec:DALAFNLabels}.


\section{Evaluation Datasets and Metrics}\label{sec:EvaluationAndDatasets}
In this section, we introduce datasets and evaluation metrics commonly used for active learning and learning with label noise. The State-of-the-art DAL methods on zero-label noise datasets are \cite{Grriz:CostEffectiveA17,Konyushkova:LearningAL17,Sener:CoreSetActiveL18,Phillips:NearCoreSets,Phillips:CoreSetsSketches16}. Datasets and their meta-data are provided that pertain to AL and DL on label noise. We conclude the section by exploring the evaluation metrics for DAL for noisy data on common bench-marking datasets.

\subsection{Datasets}\label{datasets}
As stated previously, the meteoric rise of DL algorithms was largely due to the availability of large labeled training datasets. In image classification, the most notable datasets include MNIST \cite{lecun:mnist10}, ImageNet \cite{Jia:ImageNet09}, CIFAR-100 \cite{Krizhevsky:CIFAR10009}, CALTECH-101 \cite{FeiFei:Caltech10109}, SVHN \cite{Netzer:SVHN2011}, and MS COCO \cite{Lin:MSCOCO14}. Public evaluation datasets facilitate a centralized evaluation of algorithms on a pre-defined task. These datasets can be downloaded from their websites or the different DL frameworks such as PyTorch, TensorFlow, Jax, and Theano. The best-performing models and their results on these datasets are normally hosted on a public leaderboard for the dataset. When evaluating datasets for DAL on image classification tasks, the standard practice is to use the same datasets as in full dataset image classification, but we monitor performance gain after a pre-defined number of labeled examples. While in practice this may not be the case that all labels are available upfront as is the case in using fully labeled datasets, the training cycle of DAL algorithms applied on these complete datasets substantially mimics the process of obtaining labels from an oracle for a live stream of unlabeled data within a budget.

The datasets vary widely in size, the number of classes, and the complexity inherent in telling the classes apart. Of all the commonly used AL datasets, MNIST is the least complex, with only 10 classes of hand-written digits in single channel $28\times28$ images. CALTECH-101, ImageNet, and CIFAR-100 are higher-resolution image datasets. These datasets contain more classes than MNIST, some of which are harder to tell apart. In passive learning, the model sees all available training samples per class, but in the DAL setting, depending on the query algorithm and scarcity of a class, the algorithms may never see more than a third of the samples of certain under-sampled classes. This leads to poor validation performance. 

Large-sized datasets with high-resolution images also pose a computational problem in DAL algorithms that select diverse samples based on a distance measure to all other unlabeled images. This can be extremely costly to compute in both time and hardware resource requirements. For these reasons, the reported performance of DAL algorithms on these datasets is lower than that of non-active learning algorithms since researchers have a low incentive to test complex DAL algorithms on large datasets. Table \ref{tab:datasets} contains a non-exhaustive list of commonly used image classification datasets for active learning and learning with label noise.

\begin{table}[!htb]
    \centering
    \begin{tabular}{ |p{2.5cm}|p{1cm}|p{1.8cm}|p{1.7cm}|p{3.5cm}|p{3.5cm}| }
         \hline
         Dataset & Year & \# Samples & Classes & DAL Papers & label noise Papers \\
         \hline
         \hline
         ImageNet & 2012 & 1,431,167  & 1000 & \cite{Yi:SelfSupervisedPretext22} & \cite{Hataya:InvestigatingCL18}\\
         \hline
         SVHN & 2011 & 660,000 & 10 & \cite{Gupta:NoisyBA20,Wang:BoostingALSVHN21,Sener:CoreSetActiveL18} & \cite{Gupta:NoisyBA20,Xia:PartDepLN20}\\
         \hline
         MNIST & 2010 & 70,000 & 10 & \cite{Li:Deepactivenoisestabiity22,Haubmann:DALAdaptiveAcquisition19,Gupta:NoisyBA20} & \cite{Younesian:ActiveLNoisyStrams20,Gupta:NoisyBA20}\\
         \hline
         CIFAR(10,100) & 2009 & 60,000  & (10,100) & \cite{Du:ContrastiveMismatch21,Younesian:ActiveLNoisyStrams20,Wei:NoisyAnnotation22,Li:Deepactivenoisestabiity22,Shui:DeepALUnifiedPrincipled20,Haubmann:DALAdaptiveAcquisition19,Gupta:NoisyBA20,Younesian:QActor21} & \cite{Younesian:ActiveLNoisyStrams20,Han:MaskingNoisy18,Hataya:InvestigatingCL18,Gupta:NoisyBA20,Younesian:QActor21}\\
         \hline
         Caltech101  & 2004 & 9,000  & 101  & \cite{Li:Deepactivenoisestabiity22,Yi:SelfSupervisedPretext22} & -\\
         \hline
        \end{tabular}
    \caption{Image classification datasets commonly used for deep active learning and the training of DL with noisy labels}
    \label{tab:datasets}
\end{table}

ImageNet and SVHN, being the larger of these datasets, are not well suited for DAL because training a single DL model on a large dataset is computationally expensive, and takes a lot of time. The computation complexity is worse in the case of DAL algorithms due to the iterative nature of the process. Retraining a large model over and over on the ImageNet or SVHN datasets is time-consuming. This is reflected in the literature by the reluctance of authors to use these large datasets for DAL classification, in favor of relatively small datasets such as CALTECH-101 and CIFAR-100.

Developing and training algorithms for handling noisy labels follows one of two paths: using datasets with noisy labels introduced by one or more of the noise sources listed in Section \ref{subsec:LabelNoise}, or noise-free datasets to which measurable label noise is injected by perturbing existing trusted labels. In the existing literature, the same datasets (MNIST, CIFAR-100, CALTECH-100) commonly used for image classification are used for noisy label classification, with a pre-determined probability of swapping each label. This probability is also called the noise rate, and the higher it is, the more corrupted the dataset becomes after noise injection. In literature, it is common to inject $30\%-60\%$ random symmetric label noise before training, while keeping a test set that is free of label noise. 

Datasets such as ImageNet with a large number of classes (1000) tend to also not be favored for the purpose of evaluating DAL methods that address learning under label noise. The reason here is that, with more class labels, the likelihood of class-dependent labeling errors at the time the dataset was created is higher. The kind of deliberate noise injected into datasets for noisy label learning is class-independent and is measurable as opposed to class-dependent noise that may be inherent in the data collection and annotation process. Class-dependent label noise makes training DL models harder, and there is lower confidence in the correctness of the test set labels used for evaluation. 

\subsection{Evaluation Metrics}\label{sec:evaluation}
The general evaluation methodologies for DAL algorithms on noisy labels are the same as those used for standard datasets for image classification. Top-1 accuracy is the most commonly used metric. Not much consideration is given to the underlying class distribution in most existing work and so it would be of interest to explore how class imbalances affect DAL algorithms in the presence of label noise. Since DAL is also concerned about performance under a budget, it would make sense to measure budget efficiency, a measure commonly not well documented in the literature. 

Given that active learning involves training a model a couple of times for every batch of labels received, it becomes obvious that the computational cost of DAL algorithms should be a big consideration. In  \cite{yoo:activeLCheap17}, Yoo et al. propose the use of a small network for performing query selection so that the retraining and labeling cycles run faster. Once the labeling budget is exhausted, a larger and more powerful network is then trained using the obtained labels. While this approach can be very useful in scenarios where time is of the essence, it has a big drawback. The weakness is that using a weaker learner for sample selection could lead to lower sample diversity since a weaker learner does not perform a very good job of understanding the boundaries between classes in the feature space.

In \cite{sinha:efficientDal19}, Signha et al. demonstrate the use of transfer learning for fast extraction of useful representations in a DAL setting. They show that using large pre-trained models and only fine-tuning the DAL task achieves good results with considerably fewer labeled examples. This means that, given the same budget, their approach has a higher label efficiency than a model trained from scratch. This also means for a given target performance, they require less computational resources and time to fit the target, by leveraging good pre-trained model weights. The work of Settles \cite{settles:ALSurvey09} contains a comprehensive survey of the computational cost of active learning algorithms. The main findings in this work are that the cost is influenced largely by the dataset size in terms of both the number of samples and the size of each sample. Settles also states the complexity of the query selection algorithm as well as the number of samples per batch are big factors in the total computational cost of DAL. In the case of DAL with label noise, it is critical that we have a good understanding of the underlying label noise so that the test set remains clean. While this works in developing DAL algorithms on well-known datasets, it remains unclear how the test set integrity is guaranteed in practice. If the correctness of the test labels cannot be guaranteed, evaluation methods such as top-1 accuracy, precision, and recall do not offer any reliable measure for the network's generalization performance. 

In this section, we explored datasets and evaluation metrics commonly used in comparing DAL algorithms, in particular under the setting of label noise. The next section is the main focus of this work. We explore methods leveraging the versatility of deep neural networks in the active learning framework where labeling budget is an important metric, and we are faced with a noisy label challenge.


\section{Deep Active Learning Algorithms for Noisy Labels}\label{sec:DALAFNLabels}
In this section, we focus on the main contribution of this manuscript: exploring literature on DAL algorithms used for image classification in the presence of label noise. It is worth stating that while literature is rich in theoretical approaches for handling label noise in the offline setting, very little has been done for active learning algorithms. Methods that address label noise by modeling the underlying generative distribution and filtering noisy label examples from the training set to achieve better performance are few and in between. We foresee a lot of work going into this work and look forward to understanding how iterative processes best approximate a noisy label distribution. We are also interested in understanding how low sample numbers affect label noise distributions. These ideas remain unexplored in literature. 

The methods in this section are predominantly independent of the noise distribution and seek noise-robust active training by using customized model architectures, loss functions, or training procedures. In \cite{Gupta:NoisyBA20}, Gupta et al. propose the use of standard sample diversity and importance query policies, supplemented by the model's confidence scores on samples. They argue that DNNs are normally uncertain about the decision boundaries between classes very early in training. Training with label noise exacerbates this problem since temporary and imaginary boundaries could form based on mislabeled samples, and through diversity sampling, get propagated into important query batches that influence model uncertainty and sample importance. The authors use the BALD score, introduced in \cite{Gal:BALD17} (not to be confused with the paper \cite{Cao:BALD21}) as an importance score to ensure the information content of samples for labeling per batch is optimal and a good representation of the entire dataset. The inclusion of model uncertainty in the sample selection query ensures labeled batches will include samples the current model is very uncertain about, and these, assuming satisfactory oracle label accuracy, improve the entire DAL cycle under label noise.

The inclusion of highly uncertain samples is only one-half of the novelty of their approach to robustify learning under a noisy oracle. They include a denoising layer to their network. The denoising layer is explained in the following manner: The softmax output of the original classifier is fed to the denoising layer, and the model is trained on the denoising layer, which represents a non-zero probability of predicting a particular label given the true label. This final denoising layer's weights, unlike normal final softmax outputs, are constrained to ensure noisy labels have little impact during training. During testing, the penultimate layer's output is used instead of the denoising layer for prediction. In this way, the denoising layer serves as a rigorous teacher to the student, becoming a noise-robust model used in testing and deployment.

Gupta et al. show that their method, in the noise-free setting achieves similar performance to the common baseline DAL methods, such as the original BALD \cite{Cao:BALD21}, core-set, entropy-based selection, and random sample selection on the MNIST, CIFAR10, and SVHN datasets. We attribute these results purely to the addition of model uncertainty to diversity and information gain in selecting samples. We argue the denoising layer as described in the paper would serve no purpose if the oracle provides only clean labels and so the results in the noise-free setting would be better stated as: ``no performance loss or gain" from adding the denoising layer in the noise-free setting. At $10\%$ and $30\%$ label noise, their method outperforms the reported standard benchmarks on all of the three datasets, speaking to the effectiveness of their denoising mechanism. While these are good results, they fail to demonstrate how the approach compares to similar state-of-the-art DAL methods tailored for noisy labels, and how adding the same denoising layer to DAL ResNets trained under entropy only or random sample selection compares to their approach. The paper also lacks details on the training setup that is important for reproducibility, such as the hardware used, the exact deep learning architecture, whether pre-trained weights are used or not, and the number of training epochs.

Similar to \cite{Gupta:NoisyBA20}, in \cite{Younesian:ActiveLNoisyStrams20}, Younesian et al. introduce a DAL framework (DuoLab) for training on noisy labels using weak and strong oracles. CIFAR10 and CIFAR100 are used in training and testing a CNN, with $30\%$ and $60\%$ label noise. They adopt similar criteria to \cite{Gupta:NoisyBA20} for query selection, namely using information gain and uncertainty. The weak and strong oracles refer to the innate differences in human labelers' generalization abilities and labeling quality. It is assumed that weak oracles are cheaper and more likely to produce incorrect labels than strong oracles, and so this approach's novelty is particularly more interesting in the real-world setting where there is always a need to reduce labeling fees paid to the oracles. However, the use of two oracles seems to not affect the overall performance of the DAL classifier in their work, but rather gives budget subsidies.

When it comes to dealing with noisy labels, instead of robustifying their model, Younesian et al. approach the problem by filtering out samples suspected to have noisy labels. Their DAL approach starts with an initially labeled dataset used to train the classifier, but they deviate from the conventional use of a random batch of samples to perform this initial training. Their overall approach hinges on a key and possibly flawed assumption that there exists a small and clean batch of training examples that can be used to initially train the model. While in practice we can optimistically assume it is possible to push physical data and labeling boundaries so this initial clean set is available, the paper lacks the minimum theoretical guarantees in the case we are unable to say with ``100\%" certainty that specific labels from the oracle are correct. If we were to initially assume some labels are ``100\%" correct, the paper does not make it clear as to how the correctness of such labels is verified given that the oracles have a noise rate of up to $60\%$. Once the model is trained on the initial clean samples, the model predicts the classes of all samples in the unlabeled pool, and the model's confidence score on the top 2 classes is used in deciding whether a sample is noisy or not. They measure the difference in top-2 class probabilities for each sample, and declare the top-k samples with the lowest margin as potentially noisy and so not fit for training the network. 

It becomes obvious that this approach will lead to a number of false positives and false negatives, and are likely to affect the overall performance of the model in classes that are hard to tell apart. Another issue is that a model trained on a very small subset of a large dataset can display a high top-1 class prediction confidence while being totally wrong if it has seen very little to no examples of a class in training. They combat this by reusing the noisy labels once all the clean examples are exhausted. The noisy samples are clustered based on the trained model's penultimate representation of each noisy sample. The samples within each cluster are then ranked based on their informativeness, and the top $K$ samples per cluster are picked and used to further train the model. The training batches on these samples are not random. They are ordered so that the most informative samples are in the first training batch and the least informative are in the last batch. It is unclear how this process circumvents the need for actual ground truth labels, and how this further training on noisy labels does not negatively affect the performance of a model  first trained on clean labels. Reported test accuracy results show that DuoLab outperforms noise-resilient baselines on both CIFAR 10 and 100 with $30\%$ label noise.

In \cite{Younesian:QActor21}, Younesian et al. propose QActor, an approach that follows the same idea as their earlier work in \cite{Younesian:ActiveLNoisyStrams20}: Identifying and reusing possibly noisy labels. Over and above this, the paper introduces a noise-aware informative measure, and they demonstrate for the first time how dynamic allocation of the labeling budget per query leads to better performance as compared to the convention of equal distribution of the budget across the number of query cycles. In this paper, it is assumed that a small set of clean initial training data, as well as a clean test data, are readily available. The DAL algorithm decides which samples are to be sent to the oracle for labeling. They also leave it to the DAL algorithms to decide how many of the samples fit the selection criteria and are labeled per iteration. On each iteration, a batch of highly informative samples as measured by entropy is sent to the oracle for labeling, and then the labels are compared to the current model's predictions. Samples, where there is a disagreement, are sent to a suspicious data collection. These samples are later ranked on informativeness, and resent to the oracle for relabeling, with the previously assigned label kept in mind. The authors argue that relabeling samples that the model is uncertain about and are potentially incorrectly labeled by the oracle leads to the highest gain in model performance since these are likely samples from classes that are easy to mix up.

In \cite{Huang:OracleEpiphany16}, Huang et al. approach DAL under oracle noise in a way that is different from most of the work discussed thus far. The complexity analysis performed in their work proves DAL is possible and viable with oracle epiphany. They explain that empirical evidence has shown that oracles are likely to delay providing labels on samples they are unsure about until more related examples are presented to the oracle, at which point they have an epiphany, and are able to provide a more confident label to such examples. Having had the experience of hand-labeling 1000 images from a large fisheries dataset, we agree with the authors that oracle abstention and epiphany are realistic considerations. An interesting idea mentioned in the paper is adding one more possible class label for a classification problem with $N$ classes so that the oracle has $N+1$ possible classes to choose from. This class label can be one of: "I don't know" or "unsure". Adding this choice relieves the oracle of the urge to guess between two or three most likely label assignments for a hard sample image, hence increasing the overall confidence in the correctness of the class labels that are actually provided. This is true since in a perfect world the oracle has to be the source of absolute truth, and so we are not interested in cases where the oracle guesses correctly. An oracle guessing would be useful if they are allowed to provide a measure of how confident they are of their guess.

In their approach, they use Markov chains to model when epiphany occurs for a certain class, that was previously hard to label to the oracle. The two main assumptions made in this work are: 1) The oracle is honest, and 100\% accurate on the samples he/she decides to label anything other than ``unsure". 2) Given the oracle is honest, all samples avoided will have correct labels once epiphany occurs. For both assumptions, the authors further assume once an epiphany occurs, no drastic changes in the oracle's assignment of labels will occur. While these assumptions sound reasonable, it is worth stating that in the real world, a time-constrained oracle optimizing for earnings is unlikely to avoid assigning labels to samples they are unsure about. This is especially true if they learn earlier on that samples they label "unsure" will always return, requiring more of their time. The assumption that the oracle is 100\% accurate on examples they label is also very unrealistic and does not factor in the wide-ranging spectrum of human capabilities in labeling. It would be more useful if the authors stated how 100\% oracle label accuracy would be guaranteed or tested post-epiphany. Comparing this approach to the works \cite{Younesian:QActor21,Younesian:ActiveLNoisyStrams20} of Younesian et al. a spectrum of computer-human involvement in label noise filtration can be drawn. At one end Huang et al. use the oracle as a sole decider of label uncertainty, and  in a more hybrid setting, Younesian et al. use the currently trained classifier's confidence score together with the oracle's label as a filter for potentially noisy labels. The model-only approach to noise filtration is using the confidence score margin of the top-2 predicted classes as used in \cite{Gupta:NoisyBA20}. In both Younesian et al. and Huang et al.'s approaches, the ``uncertain" samples are sent back for relabeling. 

In \cite{Amin:InducedAbstention21}, Amin et al. present the dual-purpose learning framework. While they do not explicitly focus on nor address label noise, their combined DAL, and abstention learning approaches provide valuable insight into the intricacies of DAL and abstention, which are important components that are not discussed as rigorously in \cite{Huang:OracleEpiphany16}. The two bodies of work also differ in that Amin et al. investigate the generalization bounds of the DAL and abstention setting and find interestingly that even with an unlimited labeling budget and no labeling noise, the upper bound on the observable generalization loss that exists in the passive learning case can not be guaranteed while oracle abstention is at play. The authors however do not provide empirical comparisons of their method to state-of-the-art DAL algorithms, nor do they detail how their unique dual method would impact performance when applied to such highly-performant algorithms. The work of  Yan et al. \cite{Yan:ImperfectLabelers16} is a more general approach to the work of Amin et al. since they consider DAL under imperfect labelers, and allow for abstention. Yan et al. go on to show that under strict assumptions on the dimensionality of the decision boundary, abstention, and noise rates close to the decision boundaries, their method generalizes the lower bounds on algorithms such as \cite{Amin:InducedAbstention21}. A significant contribution of their work is in demonstrating that their algorithm need not be aware of either the label noise rate or abstention rate. With restrictions on how label noise and the rate of abstention are distributed around the decision boundaries, their algorithm performs significantly better than prior methods.


\section{Conclusion and Future Research Directions}\label{sec:Conclusion}
To summarize the literature: (1) A lot of work has been done on training DL models on noisy labels in an offline setting. (2) The literature on DAL methods is also very rich in the case of no-label noise. (3) While this is an area of research that has very practical applications and financial impact, there is very little work done at the intersection of DAL and label noise.  It is worth noting that the power of ViTs has not been explored in DAL as much as it has been in fully labeled image classification datasets, and so we see huge potential in improving DAL by leveraging unique characteristics of ViTs in image classification tasks. The exploration of the transformer layers and how attention can be used in understanding diversity, importance, and uncertainty is especially intriguing to us. In Section \ref{sec:Activelearning}, works using self-supervised pre-training are presented that attain good lower-dimensional representations of the images. These have been shown to lead to a good core set of samples for labeling in an active learning framework. It is of high importance that contrastive learning methods are explored further as methods for deriving good image representations that can then be used in improving the diversity, importance, and uncertainty-based selection in queries sent to an oracle for labeling.

In terms of DAL for noisy labels, the works of Younesian et al, Huang et al. and Gupta et al. described in Section \ref{sec:DALAFNLabels} represent methods and ideas with substantial impact in DAL. In all these approaches, filtration of noise is well addressed, but with a lot of questionable assumptions. Establishing unified DAL and label noise benchmarks and datasets would add a lot of value and ensure future methods conform to realistic assumptions about the training process and the oracle. In all methods addressing this problem, only CNNs are used and the performance is only compared to baseline DAL methods as opposed to a more convincing comparison to state-of-the-art DAL methods. We would like to be able to perform the same analysis on DAL methods for label noise using ViTs against the best DAL methods. This is especially important since ViTs have been the most dominant architectural choice for image classification in recent times. It is also critical that substantial effort is put into understanding how key assumptions made in DAL methods for noisy labels affect results, and establish convergence and complexity guarantees mathematically. This means that it is not adequate to only understand DAL on noisy labels through experimentation, but also some effort needs to be put into establishing convergence guarantees at a theoretical level. 

In all existing domain literature, to the best of our knowledge, the oracle gets no feedback. It would be very interesting to explore how a cyclical feedback loop between the oracle and the model improves both of them. Intuitively we hypothesize it would help to have tips generated by the model in an unsupervised manner accessible to the oracle in cases of difficult samples. A step closer to the ideal interaction is allowing for abstention and epiphany. One way to do this is through contrastive learning. We hypothesize that, since contrastive learning models, through rotation or patch prediction for example, can be trained with no actual problem domain labels and thus noise-free, the representations learned from this step can be used in proposing similar samples to an oracle before they can abstain from labeling a sample. In this manner, the oracle may reach epiphany much earlier than they would in the frameworks listed in the literature. We would also consider answering the question of whether a real human annotator with a known or unknown noise rate can improve, and by how much in this setting.

Since reproducibility and robustness are key factors in the ongoing development of DL and AL, we are interested in performing extensive training of out-of-the-box DL models in the active learning framework in the presence of label noise. This will not only ensure researchers know what to expect out of different models without any hyperparameter tuning but also set up benchmarks that are specific to DAL with label noise. We are interested in covering as many image classification datasets as possible, using multiple CNN and ViT-based models, different AL algorithms, different noise handling algorithms, as well as different loss functions. This is vital to the field as it encourages clear and concise statements about the mode and training conditions, and ensures future methods are to be compared on a level playing field.

Lastly, a key consideration is in computation. Most work in the area does not explicitly state and discuss the computational complexity of the methods. In a world ever so gravitating towards lowering carbon footprint, it is important we are able to assess DAL methods on noisy labels not only on their accuracy but also on their computational requirement. This is an interesting avenue of research if one considers how recent work on training large language models has shown there are considerable trade-offs and gains to be made in the computational operations required, model performance, architectural choices as well as representation size. Some of these have gone in the face of established results \cite{Xiaoqi:CompBert20,Clark:ELECTRA20}.

\bibliographystyle{unsrtnat}
\bibliography{references}  

\appendix
\section*{Appendix}
\addcontentsline{toc}{section}{Appendix}


\subsection{Literature Networks}\label{sec:LiteratureTree}
Below we present a visual depiction of the literature most related to this review focusing on deep active learning on noisy labels for image classification. We present the views as images of network graphs where the nodes represent papers and the edges represent the similarity between articles. The larger the node, the more influential the article is to related articles, and the thicker the connection between any two papers, the more closely related the articles are. The Connected Papers visualization tool \cite{Eitan:ConnectedPapeers} was used to create these graphs.  

\begin{figure}[!tbp]
\centering
\adjustbox{center}{
    \includegraphics[width=\textwidth]{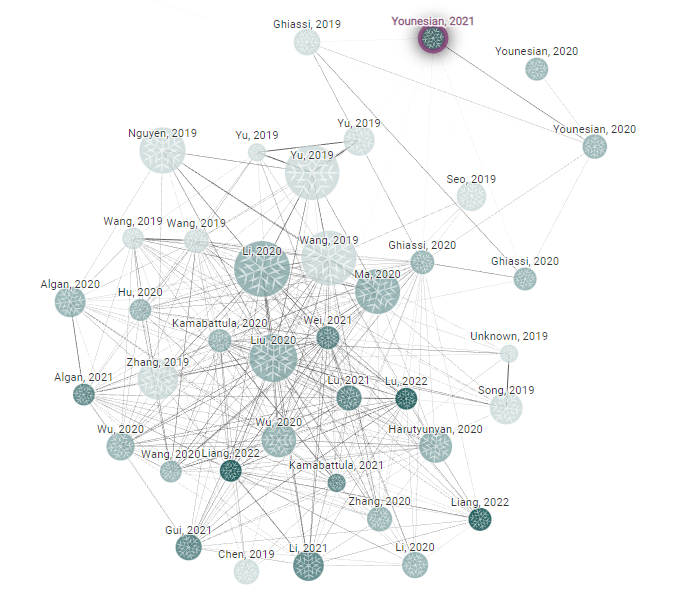}
    }
\captionof{figure}{Deep active learning with noisy labels literature closely related to Younesian et al. \cite{Younesian:QActor21}}
\label{fig:Younesian}
\end{figure}

\hspace{0.5cm}

\begin{figure}
\centering
\adjustbox{center}{
    \includegraphics[width=\textwidth]{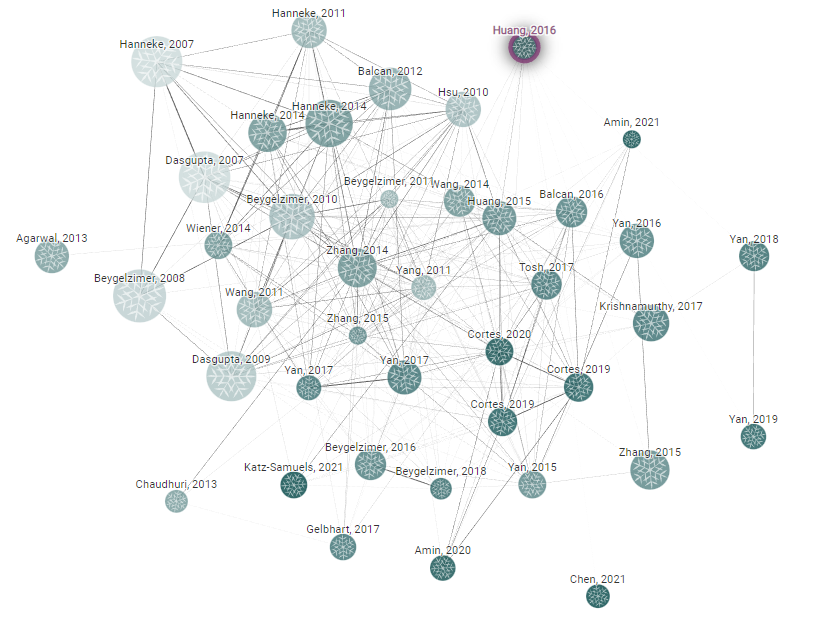}
    }
\caption{Deep active learning with noisy labels papers related to the work of Huang et al. \cite{Huang:OracleEpiphany16}}
\label{fig:huang}
\end{figure}

\hspace{0.5cm}

\begin{figure}
\centering
\adjustbox{center}{
    \includegraphics[width=\textwidth]{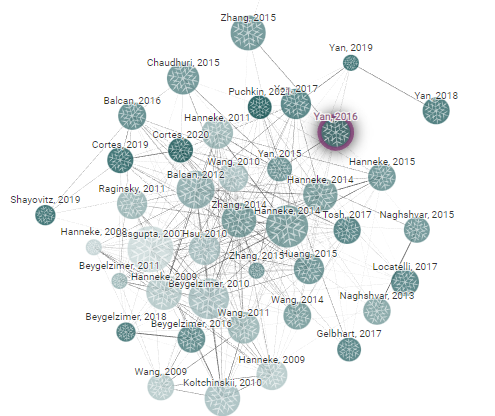}
    }
\captionof{figure}{Deep active learning with noisy labels papers related to the work of Yan et al. \cite{Yan:ImperfectLabelers16}}
\label{fig:Yan}
\end{figure}

\hspace{0.5cm}

\begin{figure}
\centering
\adjustbox{center}{
    \includegraphics[width=\textwidth]{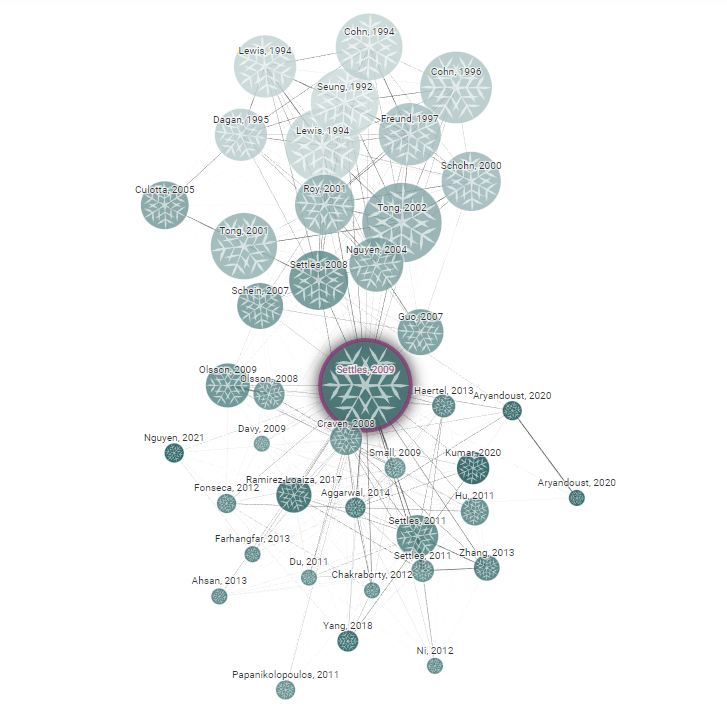}
    }
\captionof{figure}{Active learning work closely related to the survey manuscript by Settles \cite{settles:ALSurvey09}}
\label{fig:Settles}
\end{figure}

\end{document}